\documentclass[letterpaper,10pt,conference]{ieeeconf}
\IEEEoverridecommandlockouts
\usepackage{times}
\makeatletter
\let\NAT@parse\undefined
\makeatother
\usepackage[numbers,sort&compress]{natbib}
\usepackage{amsmath,amssymb}
\usepackage{graphicx}
\usepackage{capt-of}
\usepackage{textcomp}
\usepackage{xcolor}
\usepackage{siunitx}
\DeclareSIUnit{\byte}{B}
\usepackage{booktabs}
\usepackage{url}
\usepackage{stfloats}
\usepackage{balance}
\usepackage[hidelinks]{hyperref}
\graphicspath{{Figures/}}

\title{\LARGE\bf Around the World: Unified Learned Locomotion on a\\
270\,g Continuous-Rotation Quadruped}
\author{Arturo Flores Alvarez, Nathan Lintu, and Dennis Hong%
\thanks{All authors are with the Robotics and Mechanisms Laboratory
(RoMeLa), University of California, Los Angeles (UCLA).\protect\newline
\texttt{\{afloresa,nlintu,dennishong\}@g.ucla.edu}}}

\IEEEaftertitletext{% Full-width first-page teaser, placed by the conference class's title hook.
\begin{minipage}{\textwidth}
  \centering
  \includegraphics[width=\textwidth]{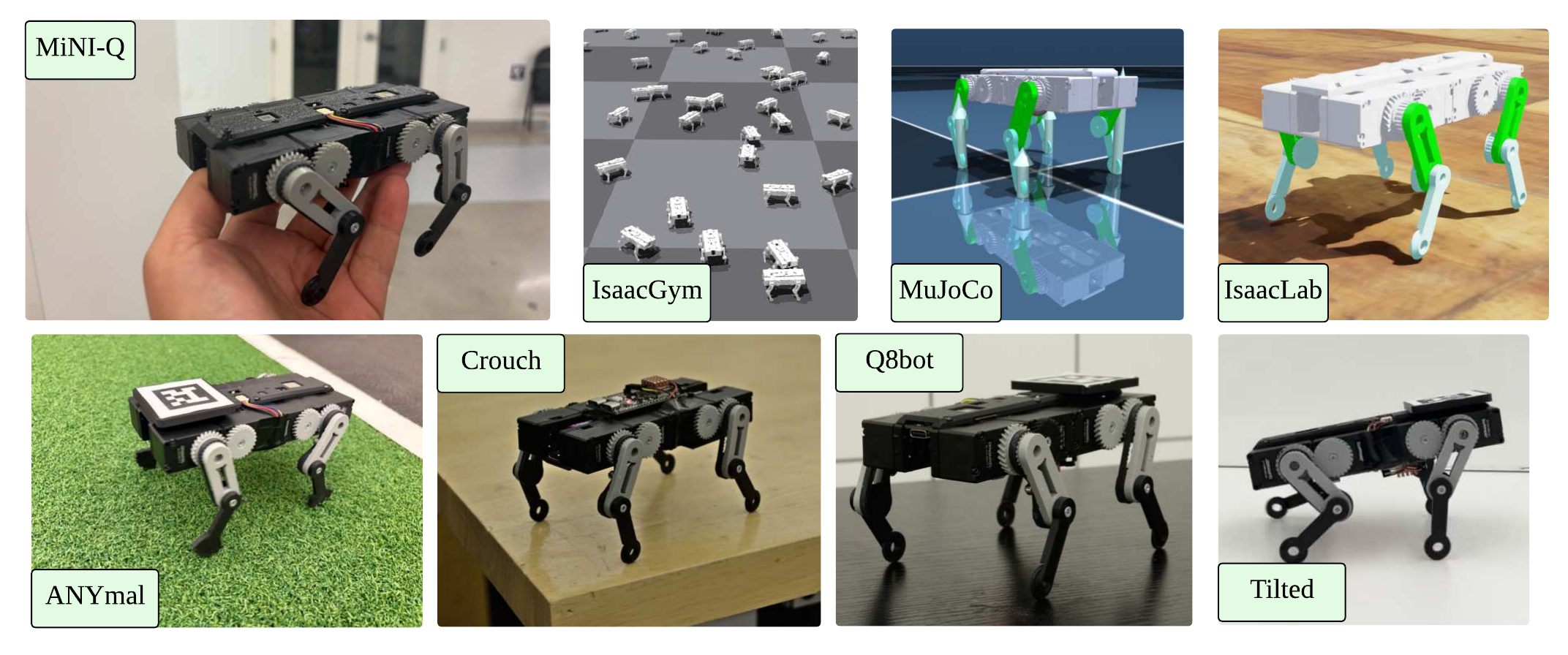}
  \captionof{figure}{Overview of the MiNI-Q pipeline and commanded
  posture family. A single policy is trained in massively parallel
  simulation, validated across physics engines, and deployed on the
  \SI{270}{\gram} robot. Continuous-rotation legs and a
  gravity-conditioned reference extend the same onboard control
  interface to upright walking, inverted walking, and landing recovery.}
  \label{fig:pose_grid}
  \medskip
\end{minipage}
}

\def\mainpaper{1}
\begin{document}
% Pass formatting control to BibTeX without creating a numbered citation.
\makeatletter
\immediate\write\@auxout{\string\citation{miniq:bst-control}}
\makeatother
\maketitle
\thispagestyle{empty}
\pagestyle{empty}

\begin{abstract}
Closed-loop learned locomotion is established on commercial quadrupeds
but remains uncommon at the sub-kilogram scale. Continuous-rotation
legs give MiNI-Q, a \SI{270}{\gram} quadruped, access to supporting
configurations on either side of the body. We exploit this range with
a single posture-conditioned reinforcement-learning policy that runs
entirely onboard. A continuous joint-space reference on the torus
$T^8$ and its gravity-conditioned transformation connect upright
walking, inverted walking, and landing recovery without state machines
or phase switching. Coordinated posture and release curricula train
this behavior family; identified actuation, cross-engine validation,
and embedded execution support hardware transfer. The same
sub-100k-parameter network tracks forward velocity with RMSE of
\SI{0.037}{\meter\per\second} upright and
\SI{0.050}{\meter\per\second} inverted, resumes walking in 24 of 30
release trials, and operates with four interchangeable foot geometries
across four indoor surfaces. Hardware experiments and simulation
ablations connect these capabilities to the representation,
conditioning, and training choices that exploit the platform's motion
range. Demonstration videos and supplementary material are available
on the project website:
\url{https://submissionreview.github.io/around-the-world/}.
\end{abstract}


% Queue the reference-family figure for the next full-width page slot.
% Paired posture-reference figure; kept separate to tune float placement.
\begin{figure*}[t]
  \centering
  \begin{minipage}[c]{0.48\textwidth}
    \includegraphics[width=\linewidth]{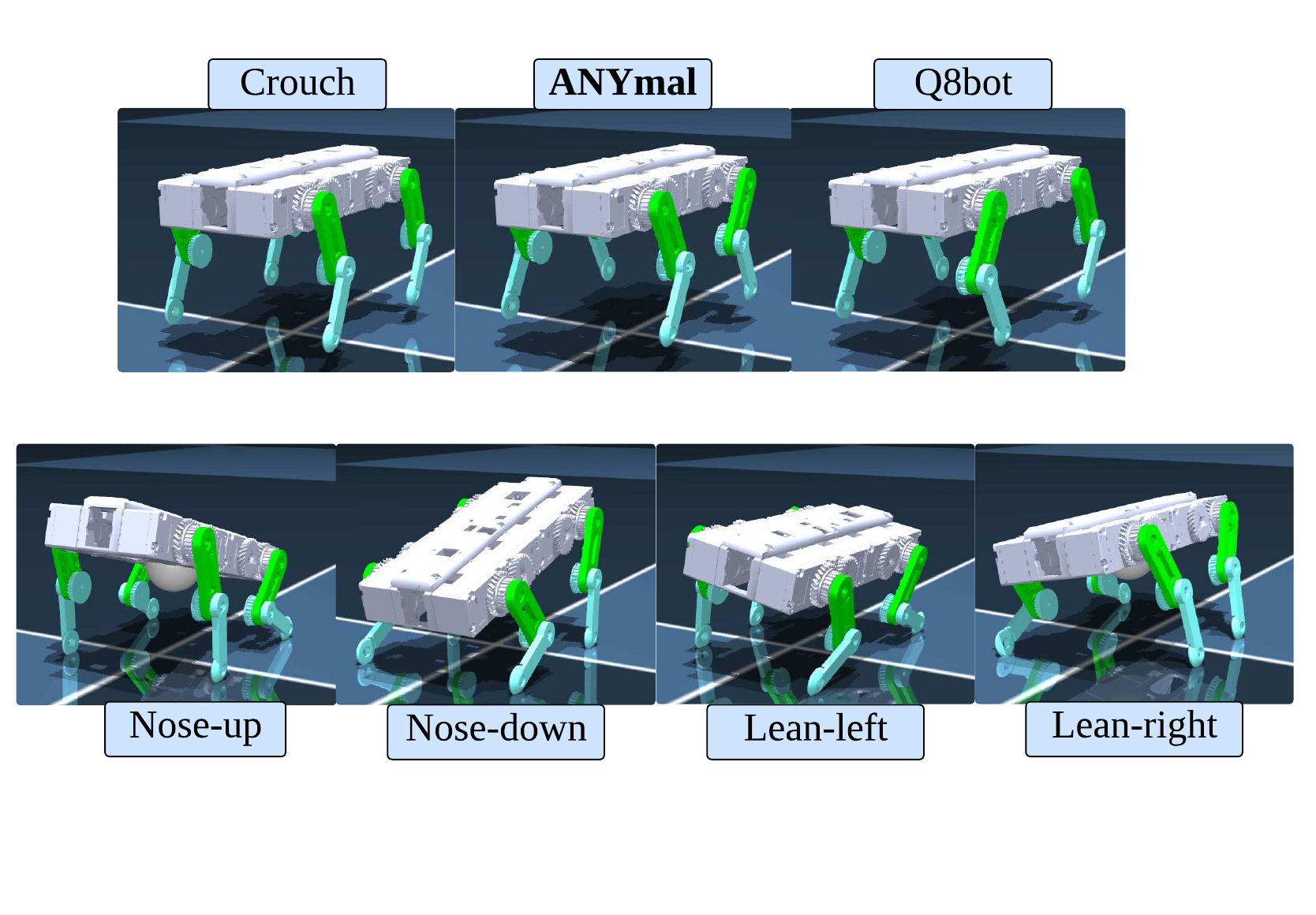}
  \end{minipage}\hfill
  \begin{minipage}[c]{0.50\textwidth}
    \includegraphics[width=\linewidth]{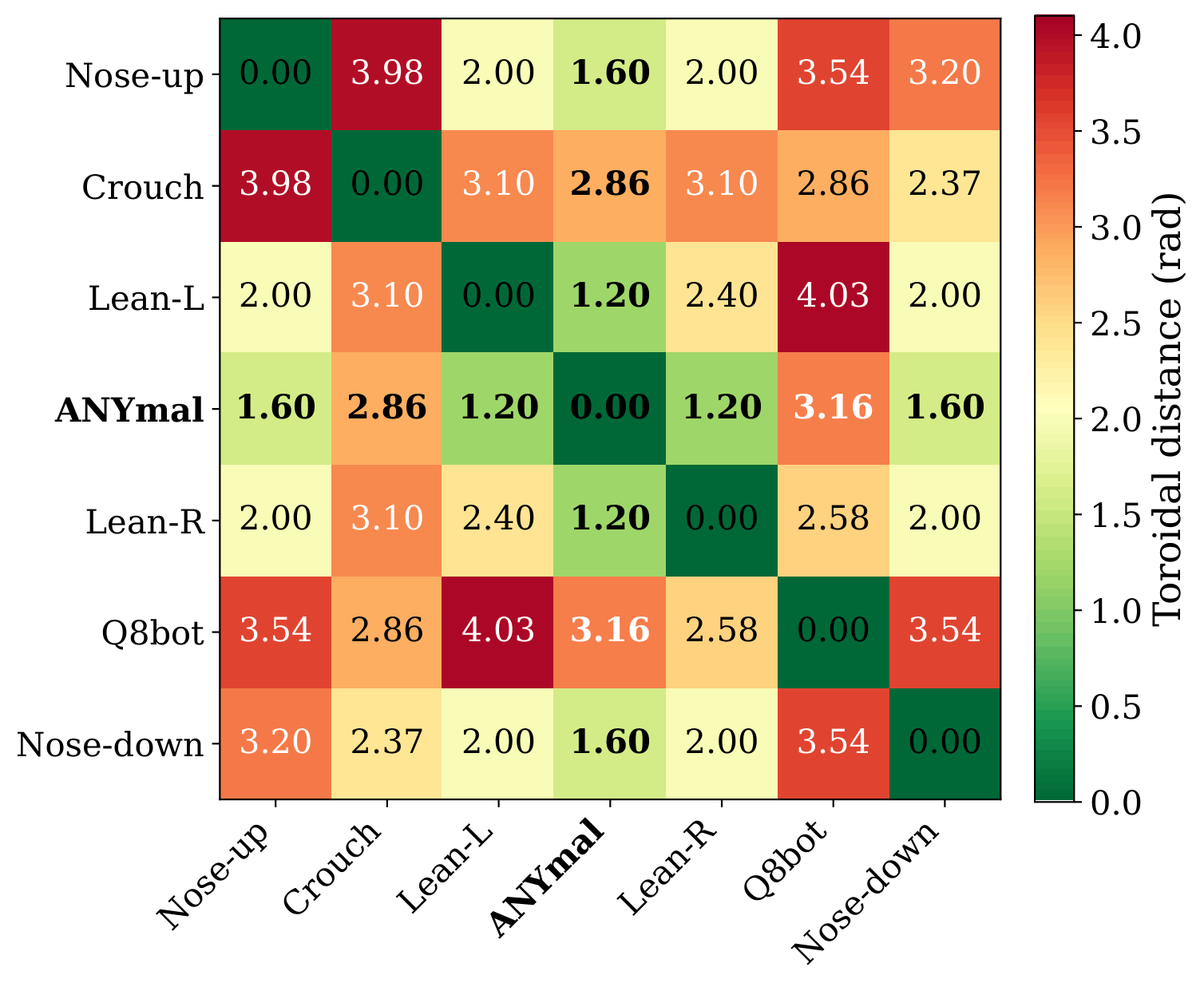}
  \end{minipage}
  \caption{Posture references and curriculum geometry. Left: three stance
  configurations (top) and four attitude references around ANYmal
  (bottom), all on MiNI-Q. Right: pairwise toroidal distances (rad).
  Lean-left--Q8bot is the only pair beyond the stage-2 cutoff of
  \SI{4.0}{\radian}. Training samples both anchors and their interpolations.}
  \label{fig:anchors}
  \label{fig:anchor_heatmap}
\end{figure*}

\section{Introduction}
\label{sec:introduction}

Reinforcement learning has enabled robust locomotion and diverse
behaviors on quadruped robots~\citep{hwangbo2019anymal,rudin2022walk,margolis2022walkthese}.
At the sub-kilogram scale, learned locomotion remains less
established~\citep{jabbour2022closing,akgun2024tinyrl}. Small, low-cost
quadrupeds are widely available for educational and research use, but
closed-loop learned control with sim-to-real transfer at this mass is
comparatively rare. The closest sibling platforms in this class,
MiNI-Q~\citep{koh2026miniq} and Q8bot~\citep{wu2025q8bot}, have been
reported as hardware introductions or under open-loop control.
MiNI-Q is a \SI{270}{\gram} quadruped whose eight belt-driven joints
rotate without mechanical stops. This range permits supporting leg
configurations on either side of the body and continuous changes of
stance.

That topology has two consequences for control. First, each joint
lives on the circle $S^1$ rather than a bounded interval, so the joint
space is the torus $T^8=(S^1)^8$. One policy can therefore represent
inverted gaits, mid-flight repositioning, and interpolated postures
that joint limits would exclude. Second, the platform is modular
enough that reprinting lower links, swapping foot geometries, and
reflashing compute are part of ordinary laboratory iteration. These
properties also introduce specific challenges. Unbounded joints break
standard observation pipelines at $\pm\pi$. The \SI{270}{\gram} mass
sits near simulator integration regimes where effort-mode and
position-mode actuation diverge, and the lower links weigh only
\SI{1}{\gram} each. The belt drive introduces measurable backlash, the
onboard compute is a pair of microcontrollers linked over
I\textsuperscript{2}C rather than a single processor, and printed foot
geometry varies with manufacturing tolerance and wear.

On this miniature platform, learning must turn the continuous-rotation
range into a controller that follows posture and velocity commands,
repositions its legs during a fall, and resumes walking after contact.
Periodic joint representations and training across postures and
release conditions connect these behaviors. Hardware deployment
further requires transfer to the dynamics of the geared servos,
lightweight links, and interchangeable feet within the robot's onboard
sensing and compute budget.

The published MiNI-Q platform demonstrates inverted mobility using
precomputed trajectories~\citep{koh2026miniq}. We exploit its leg
symmetry to connect these behaviors in one closed-loop policy. To our
knowledge, this is the first demonstration on a sub-kilogram quadruped
of a single learned policy, executed entirely on onboard
microcontrollers, that unifies upright and inverted locomotion,
continuous posture conditioning, and landing recovery.
Our contributions are:
\begin{enumerate}
  \item \textbf{Unified locomotion and recovery across body orientations.}
  A gravity-conditioned reference exploits the leg symmetry to connect
  upright walking, inverted walking, and landing recovery in one
  policy, with periodic observations and a release curriculum. No
  state machines or phase switching are used; a compact network runs
  entirely onboard at \SI{50}{\hertz}.
  \item \textbf{Continuous posture conditioning during locomotion.}
  A continuous joint-space command on $T^8$ represents stance, body
  attitude, and interpolated references, including their inverted
  counterparts, while the same policy follows a velocity command. An
  interpolation curriculum reaches intermediate postures that are
  never presented as discrete training targets.
  \item \begin{samepage}\textbf{Onboard transfer across foot geometries.}
  Foot-shape randomization, identified actuation, and embedded execution
  support the same trained policy across tested foot variants and
  surfaces without retraining, connecting the formulation to a modular
  miniature robot.\par\end{samepage}
\end{enumerate}

Each contribution is evaluated in Section~\ref{sec:experiments} and
ablated in Section~\ref{sec:ablations}. The supporting identification,
validation, and onboard inference pipeline is detailed in
Sections~\ref{sec:method}--\ref{sec:system} and
Appendix~\ref{app:engineering}.

\section{Related Work}
\label{sec:related_work}

\paragraph{Compact platforms and embedded learned control}
MiNI-Q~\citep{koh2026miniq} and Q8bot~\citep{wu2025q8bot} explore
miniature quadruped mechanisms and electronics. Stanford
Doggo~\citep{kau2019stanforddoggo} and Pupper~\citep{kau2022pupper}
provide accessible platforms for dynamic locomotion and benchmarking.
Jabbour et al.~\citep{jabbour2022closing} deploy learned walking on
Petoi Bittle using an onboard Raspberry Pi Zero 2 W. TinyRL studies
compressed Bittle policies in simulation~\citep{akgun2024tinyrl};
learned source seeking has been
executed on a nano-drone microcontroller~\citep{duisterhof2019learning}.
RLtools~\citep{eschmann2024rltools}, ARM-oriented
CMSIS-NN~\citep{lai2018cmsisnn}, and related onboard inference
tooling~\citep{pavel2026onboard} address efficient learning and neural
inference under embedded resource limits. Closed-loop learned control
on sub-kilogram quadrupeds remains uncommon; most published results
report open-loop motion replay or off-board policy execution. We study
continuous-rotation behaviors with the entire policy running on a
dual-microcontroller stack and no off-board compute in the loop.

\paragraph{Locomotion learning and sim-to-real transfer}
Massively parallel simulation and proximal policy
optimization~\citep{schulman2017ppo,makoviychuk2021isaac,mittal2023orbit,todorov2012mujoco,rudin2022walk}
have made data-efficient training of locomotion policies routine.
Learning and optimization have enabled quadruped locomotion over
challenging terrain~\citep{kalakrishnan2011learning,grandia2023perceptive}.
Actuator and contact modeling~\citep{tan2018sim2real,hwangbo2019anymal,hwangbo2018per}
support sim-to-real reinforcement learning~\citep{lee2020learning},
with demonstrations on bipeds as well as
quadrupeds~\citep{xie2020learning,siekmann2021blind}.
Energy-based reward shaping~\citep{fu2021energy} and concurrent state
estimation~\citep{ji2022concurrent} address gait emergence and
imperfect observations. Rapid motor adaptation~\citep{kumar2021rma}
and domain randomization~\citep{tobin2017domain,peng2018sim}
also support transfer in dexterous
manipulation~\citep{andrychowicz2020learning,handa2023dextreme}.
Perception-aware locomotion~\citep{miki2022learning,loquercio2021learning}
and recent surveys~\citep{fan2024review} cover the heavier mass class.
We reuse standard learning components and apply identified servo
dynamics and randomized feet to transfer the posture and recovery
formulation at sub-kilogram scale.

\paragraph{Behavior conditioning, agility, and recovery}
Walk These Ways~\citep{margolis2022walkthese} and
AllGaits~\citep{bellegarda2024allgaits} encode families of locomotion
behaviors, while ManyQuadrupeds~\citep{shafiee2023manyquadrupeds}
studies one policy across robot morphologies. Learned locomotion also
supports rapid running~\citep{margolis2022rapid}, agility
benchmarks~\citep{caluwaerts2023barkour}, and
parkour~\citep{cheng2024extreme}. Perceptive systems extend the motion
repertoire through terrain-aware walking~\citep{miki2022learning} and
visually guided jumping~\citep{margolis2023learning}. Hierarchical or
recovery-specific controllers such as ANYmal
recovery~\citep{lee2019recovery} target the fall--recover--walk regime
with separate behavior policies. We study one posture-conditioned
policy that resumes walking in either body orientation after landing,
without a separate recovery controller.

\paragraph{Posture commands and continuous rotation}
Those behavior-family methods operate on bounded-joint quadrupeds and
constrain the command set to the kinematically reachable subspace.
Conditioning on $T^8$ is a natural fit for a platform whose joints are
unbounded. Wheeled-legged robots exploit continuous actuator rotation
for hybrid locomotion~\citep{bjelonic2020wheellegged}, and the
published MiNI-Q platform demonstrates inverted mobility using
precomputed trajectories~\citep{koh2026miniq}. Closely related
miniature quadrupeds such as Q8bot~\citep{wu2025q8bot} and Stanford
Pupper~\citep{kau2022pupper} retain conventional bounded revolute
joints. Here, an eight-joint posture reference specifies stance and
body attitude on a fixed robot. Its periodic representation and
gravity-conditioned transformation let the same command interface
describe upright and inverted configurations. Continuous rotation
admits references that joint limits can exclude, while conditioning
represents them within one policy. The foot-swap study then tests the
controller under physical changes to contact geometry.

\section{Learning Locomotion and Recovery}
\label{sec:method}

\subsection{Task and Command Space}
\label{sec:method_policy}
Building on established reinforcement-learning approaches to quadruped
locomotion~\citep{hwangbo2019anymal,rudin2022walk}, we train one policy to
track velocity and posture commands and resume walking after landing.
On MiNI-Q, upright and
inverted supporting configurations are mechanically accessible, with
contact and collision constraints determining locomotion feasibility.
Commands specify forward velocity, yaw rate, and a continuous
joint-space posture $\theta^p$. The policy produces eight offsets
about the selected reference and remains active during flight,
landing, and walking. There is no state-machine layer between these
regimes: the same network runs at every step. The actor is a
$[128,128]$ ELU MLP sized for onboard headroom rather than to match
the larger networks used on heavier quadrupeds; training settings and
the privileged critic observation are given in
Section~\ref{sec:exp_setup} and Appendix~\ref{app:priv_obs}.
Actions are eight joint-position offsets added to the selected
reference and tracked by the identified position drive.

\subsection{Posture Conditioning and Orientation Symmetry}
\label{sec:method_morphology}
Figure~\ref{fig:anchors} shows seven reference anchors: Crouch,
ANYmal, Q8bot, Nose-up, Nose-down, Lean-left, and Lean-right.
The first three specify stance configurations on MiNI-Q; the other
four vary body attitude around the ANYmal reference. Encoding a target
as joint angles permits both these anchors and intermediate postures
to condition the same policy. Each joint angle lies on the circle
$S^1$, identifying angles that differ by $2\pi$. Their eightfold
Cartesian product, $T^8=(S^1)^8$, is the joint torus. Distances between
configurations $\theta,\theta'$ use shortest wrapped angular differences:
\begin{equation}
 d_{T^8}(\theta,\theta')=
 \sqrt{\sum_{i=1}^{8}\mathrm{wrap}(\theta_i-\theta'_i)^2},
 \quad \mathrm{wrap}(\cdot)\in(-\pi,\pi].
 \label{eq:dT8}
\end{equation}
For anchors $\theta^A,\theta^B$, interpolation follows the wrapped
joint differences:
\begin{equation}
 \theta^p(u)=\mathrm{wrap}\!\left(\theta^A+
 u\,\mathrm{wrap}(\theta^B-\theta^A)\right),\quad u\in[0,1].
 \label{eq:interpolation}
\end{equation}
The distance also defines the reference-pose penalty. Planar forward
kinematics maps the reference to the projected-gravity target used by
the orientation reward, so the tilted anchors specify body attitude
through the same objective. Attitude control therefore emerges as a
derived capability of the posture family rather than a separate
task (Appendix~\ref{app:attitude_control}). The interpolation curriculum
reaches geodesic intermediates that never appear as discrete training
targets (Appendix~\ref{app:anchor_selection}).

\label{sec:method_z2}
For each planar leg, $\mathcal{I}(a,b)=(\pi-a,-b)$ maps the two joint
angles to the corresponding inverted configuration, modulo $2\pi$.
The inverted pose places the foot at the same body-frame position with
the robot rotated by \SI{180}{\degree} about the roll axis. Applying
this transformation twice returns the original reference.
The sign of body-frame projected gravity selects the reference:
\begin{equation}
 \theta^{\mathrm{ref}}=
 \begin{cases}
   \theta^p, & g_z<0\quad\text{(upright)},\\
   \mathcal{I}(\theta^p), & g_z>0\quad\text{(inverted)}.
 \end{cases}
 \label{eq:reference}
\end{equation}
Continuous rotation gives access to the transformed anchor or
interpolated posture, and joint-space conditioning represents that
target directly. The same mapping would be unsound on a platform whose
joint limits exclude the antipodal pose. The yaw convention uses
$s=-\mathrm{sign}(g_z)$ to preserve command interpretation across
orientations. Reference selection changes the target algebraically
without selecting a separate recovery controller.

% Queue the control diagram near the transition to the system section.
% Original author diagram, retained as a single-column PDF figure.
\begin{figure}[t]
  \centering
  \includegraphics[width=\columnwidth]{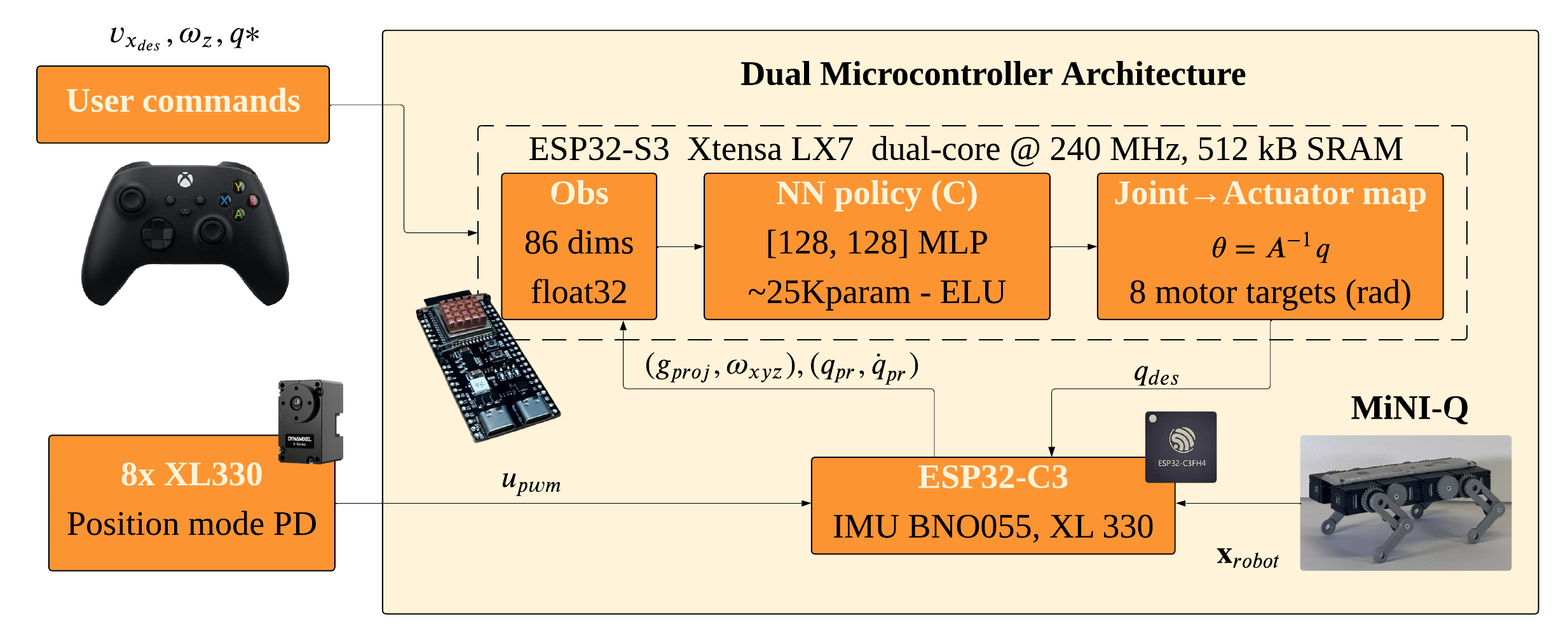}
  \caption{Onboard data path at \SI{50}{\hertz}. The S3 constructs the
  reference and observation and maps policy outputs to position targets;
  the C3 services the IMU and actuator bus.}
  \label{fig:realtime}
\end{figure}

\subsection{Periodic Observations and Policy Outputs}
\label{sec:method_toroidal}
A policy that observes raw joint angles on $S^1$ sees a discontinuity
at $\pm\pi$: two physically identical poses generate observations that
differ by $2\pi$. To retain the correspondence across that cut, we
encode angles with $E(\theta)=(\sin\theta,\cos\theta)$, applied to both
joint offsets and the reference. The map is the Whitney embedding
$S^1\hookrightarrow\mathbb{R}^2$; applied per joint, it lifts to a
bounded-distortion embedding of $T^8$
(Appendix~\ref{app:toroidal_math}). A three-frame history of joint
offsets and projected gravity supplies short-horizon motion
information without a direct joint-velocity input. The policy observes
\begin{equation}
\begin{aligned}
 o_t = [&\underbrace{\omega_b}_{3},\;
 \underbrace{g_{t:t-2}}_{9},\;
 \underbrace{E(\Delta\theta_{t:t-2})}_{48},\\
 &\underbrace{v_x^{\mathrm{cmd}},\omega_z^{\mathrm{cmd}}}_{2},\;
 \underbrace{E(\theta^{\mathrm{ref}})}_{16},\;
 \underbrace{a_{t-1}}_{8}]\in\mathbb{R}^{86},
\end{aligned}
\label{eq:observation}
\end{equation}
where $\omega_b$ is body angular velocity, $g$ is projected gravity,
$\Delta\theta=\theta-\theta^{\mathrm{ref}}$, and $a_{t-1}$ is the
previous action. All channels are available onboard. The history
channels give implicit access to short-horizon velocity without
exposing the noisy onboard velocity estimate. The eight action
offsets are added to the selected reference and sent as position
targets. PD targets are wrapped to $(-\pi,\pi]$ at every step, so a
leg can complete a full rotation across episode boundaries without a
sign flip in the commanded torque.

\subsection{Recovery and Posture Curricula}
\label{sec:method_training}
\label{sec:method_throw}
We broaden the posture and release distributions together
(Table~\ref{tab:curriculum}) to train both command transitions and
recovery. Posture sampling progresses from anchors with jitter to
nearby-pair interpolations and then arbitrary anchor pairs. Resampling
targets every \SI{6}{\second}, including within episodes, trains
transitions during locomotion.

The nearby-pair cutoff is \SI{4.0}{\radian} under Eq.~\eqref{eq:dT8}.
In Figure~\ref{fig:anchor_heatmap}, the attitude anchors lie within
\SI{1.6}{\radian} of ANYmal, while Crouch and Q8bot are
\SI{2.86}{\radian} and \SI{3.16}{\radian} away. Only
Lean-left--Q8bot (\SI{4.03}{\radian}) enters at the final stage.
This selection limits the simultaneous introduction of distant pairs,
which destabilized preliminary training attempts.

The release curriculum progresses from upright starts to lateral drops
and then orientations sampled over $\mathrm{SO}(3)$, at heights up to
\SI{0.5}{\meter}. Spawn degrees of freedom are set to the current
reference regardless of orientation, so an inverted spawn can leave
the feet directed away from the ground and require mid-air
repositioning before contact. The same policy must then establish
support and follow the locomotion command. The objective combines
velocity and yaw tracking with reference-posture, attitude, contact,
and effort terms, following locomotion reward-shaping
practice~\citep{rudin2022walk,fu2021energy}.
Table~\ref{tab:rewards} records the complete reward configuration.

\begin{table}[t]
 \centering
 \caption{Joint release/posture curriculum; reference resampling every
 \SI{6}{\second}.}
 \label{tab:curriculum}
 \footnotesize
 \setlength{\tabcolsep}{3pt}
 \begin{tabular}{@{}lll@{}}
 \toprule
 Iterations & Release conditions & Posture sampling \\
 \midrule
 0--5k & Upright, small perturbations & Anchor + jitter \\
 5k--10k & Lateral drops, $\leq$\SI{0.2}{\meter} & Pairs $d_{T^8}\leq4.0$ \\
 10k--20k & $\mathrm{SO}(3)$, $\leq$\SI{0.5}{\meter} & Any anchor pair \\
 \bottomrule
 \end{tabular}
\end{table}

\subsection{Foot-Shape Randomization}
\label{sec:method_footdr}
Replaceable feet motivate training across contact geometries through
domain randomization~\citep{tobin2017domain,peng2018sim}.
The physical lower-link tip is a printed cap whose effective contact
diameter ranges from approximately \SIrange{12}{16}{\milli\meter}
under manufacturing tolerance and wear of the elastic insert. Four
preloaded assets comprise a sphere of radius
\SI{7.5}{\milli\meter} and cylinders of length \SI{2}{\milli\meter},
\SI{3.5}{\milli\meter}, and \SI{5}{\milli\meter}. Each environment
receives one asset at initialization and a \SIrange{0}{2}{\milli\meter}
contact-distance offset representing variation in the replaceable
insert. The policy receives no foot-shape identifier. Hardware swaps
test transfer to additional printed geometries
(Section~\ref{sec:exp_foot}).

% Queue the tracking plot for the page where evaluation begins.
% Hardware tracking figure, queued near the start of the evaluation.
\begin{figure*}[t]
 \centering
 \includegraphics[width=\textwidth]{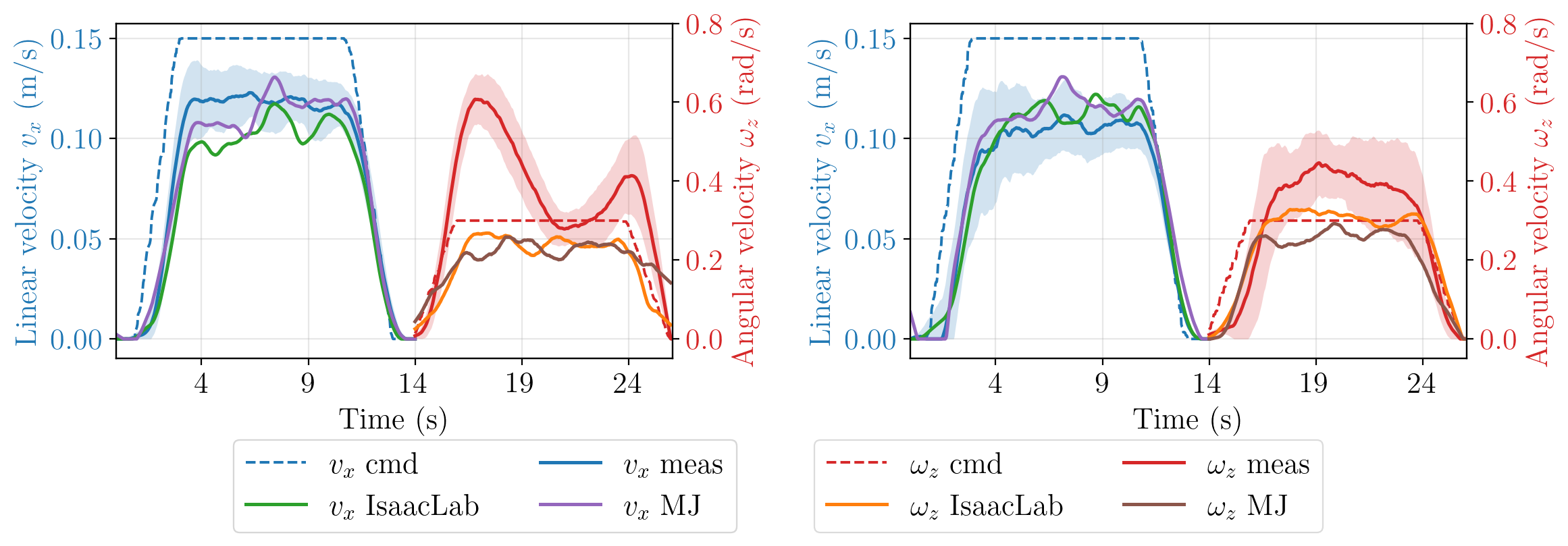}
 \caption{Velocity tracking with the same policy, upright (left) and
 inverted (right), on the real robot, Isaac Lab, and MuJoCo. Forward
 velocity and yaw rate follow a ramp-and-hold command. Dashed lines
 denote commands, solid lines measured motion, and bands $\pm1\sigma$.
 Hardware trials use $N=5$ per orientation.}
 \label{fig:tracking}
\end{figure*}

\section{System Overview}
\label{sec:system}

\subsection{Platform and Control Interface}
\label{sec:sys_platform}
MiNI-Q~\citep{koh2026miniq}, shown in Figure~\ref{fig:pose_grid},
is a \SI{270}{\gram} quadruped with a smartphone-sized footprint
for laboratory and desktop evaluation. It has four planar,
two-degree-of-freedom legs actuated by eight Dynamixel XL330-M077-T
servos (TTL bus, gear ratio $\sim 353.5$, stall torque
$\sim$\SI{0.18}{\newton\meter}) through parallel belt drives.
The platform has replaceable printed lower links and no
hip abduction--adduction joint. The nominal lower-link tip is a
3D-printed cap with an approximately spherical contact geometry of
\SI{7.5}{\milli\meter} radius; the replaceable tip and elastic insert
make contact geometry a practical source of variation. The belt
transmission gives access to the continuous reference family in
Section~\ref{sec:method} and removes the mechanical end-stops typical
of revolute-joint quadrupeds, while measurable backlash and
approximately \SI{1}{\gram} lower links make actuator fidelity relevant
to transferring the learned behavior.

The actuator firmware interface is position-only: the controller
writes a target, and the internal proportional--derivative loop
synthesizes a pulse-width-modulated command. There is no externally
writable torque command; the effort-mode identification in
Section~\ref{sec:sys_sysid} is recovered only inside the simulator.
The policy sends joint-position targets at \SI{50}{\hertz}, tracked by
the actuator's internal controller. Joint states, reference offsets,
and position targets use consistent angle wrapping across training
and firmware.

\subsection{Hardware-Calibrated Actuation}
\label{sec:sys_sysid}
\label{sec:method_sysid}
We identify the effective response of MiNI-Q's geared servos and
lightweight leg assembly to model the actuation encountered by the
deployed policy. Identification uses a single leg on a floating
fixture, with the base free along gravity and the foot unconstrained.
Each joint is excited with step references and a
sinusoidal sweep at the firmware gains $P=D=400$. The actuator model
uses saturated effort commands at \SI{1}{\kilo\hertz}.
Nelder--Mead fitting minimizes joint-position and velocity trajectory
error over stiffness,
damping, armature, and friction. Constraining stiffness near the
firmware nominal reduces its ambiguity with armature under saturation.
Appendix~\ref{app:sysid_details} gives the identification procedure.

The fitted parameters are transferred to position-mode training in
Isaac Gym at $\mathrm{d}t=\SI{2}{\milli\second}$. The XL330 supports
multiple firmware control modes; we deploy in position mode on the
robot. Trained trajectories are replayed in MuJoCo and Isaac Lab
before hardware deployment (Appendix~\ref{app:simulation}). The
fitted model is shared by training and evaluation;
Section~\ref{sec:exp_system} reports its measured fidelity.

\subsection{Embedded Execution}
\label{sec:sys_onboard}
\label{sec:method_dualmcu}
Figure~\ref{fig:realtime} shows the onboard data path. An ESP32-S3
(dual-core Xtensa LX7 at \SI{240}{\mega\hertz}, with SIMD via PIE128)
runs the compiled policy as a high-priority FreeRTOS task on one core;
its second core services communication with an ESP32-C3 over
I\textsuperscript{2}C. The C3 handles the Dynamixel TTL bus and BNO055
IMU. The policy uses proprioceptive observations and user commands;
external tracking is used only for evaluation. The loop is
synchronous at \SI{50}{\hertz} with no off-board compute and no
wireless link in the control path.

The compiled $[128,128]$ policy uses the S3's PIE128 SIMD instructions
with weights stored in DRAM, leaving time for observation construction
and peripheral exchange within the \SI{20}{\milli\second} period.
Table~\ref{tab:nn_perf} reports the measured inference cost.
Appendix~\ref{app:dualmcu} details the onboard implementation.

\section{Experimental Setup}
\label{sec:exp_setup}

\subsection{Training Configuration}
We train a $[128,128]$ ELU actor and critic with
PPO~\citep{schulman2017ppo} in 4096 parallel Isaac Gym
environments~\citep{makoviychuk2021isaac,rudin2022walk}, using a
\SI{2}{\milli\second} physics step and a \SI{50}{\hertz} policy rate.
The critic additionally receives 22 privileged channels describing
base motion and height, disturbances, noise-free joint velocities,
and mass properties; these are excluded from the deployed policy
(Appendix~\ref{app:priv_obs}).
PPO uses 24-step rollouts, 20 mini-epochs, discount $0.99$, GAE
parameter $0.95$, adaptive learning rate with KL target approximately
$0.01$, and entropy coefficient $-0.015$ in the minimized loss.
The 20{,}000-iteration schedule takes approximately eight hours on one
NVIDIA RTX 4090 at roughly 70k environment steps per second
(approximately 100k steps per second for baseline walking without the
full feature set).

Table~\ref{tab:rewards} gives the reward terms and coefficients.
Here $c_k$ indicates foot contact, $t_k^{\mathrm{air}}$ is swing duration,
and $\hat f_k$ is the foot direction in the gravity-alignment term.
The posture reference determines both the joint-space and attitude
targets, so the same objective applies across the command family.
Appendix~\ref{app:hyperparams} records the training hyperparameters
and supporting training diagnostics.

% Queue before the reward table to place the posture figure first in its column.
\begin{figure}[t]
 \centering
 \includegraphics[width=\columnwidth]{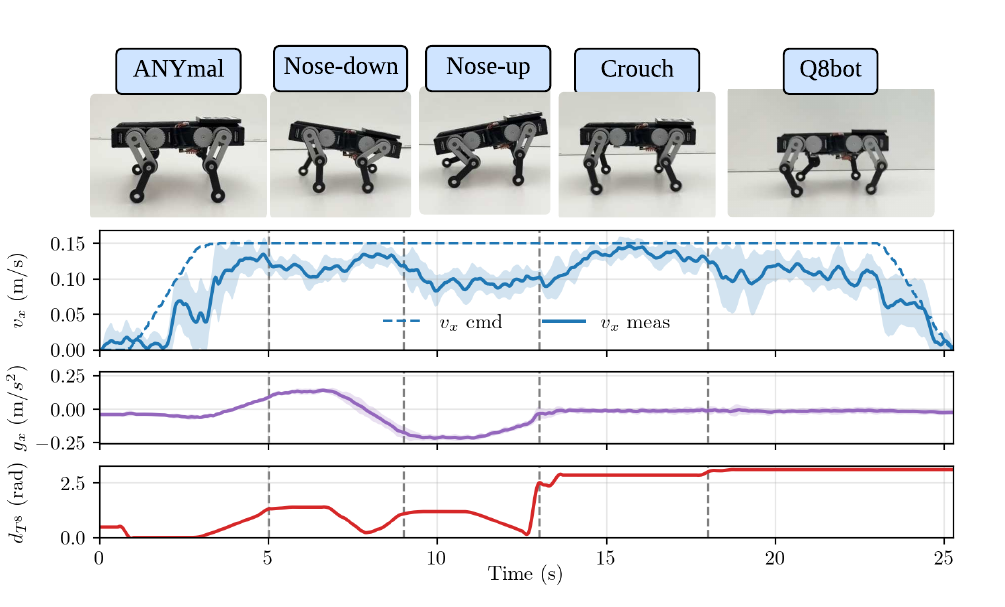}
 \caption{Hardware posture changes under a continuous velocity command
 (\SI{25}{\second}, $N=5$). Plots report $v_x$, $\omega_z$, and
 toroidal distance $d_{T^8}$ from the current joint state to the fixed
 ANYmal anchor; the images show the commanded posture sequence.}
 \label{fig:pose_tracking}
\end{figure}

% Shared reward table. Expressions and coefficients match the CoRL source.
\begin{table}[t]
  \centering
  \caption{Reward configuration. $e_v=v_x^{\mathrm{cmd}}-v_x$,
  $e_\omega=\omega_z^{\mathrm{cmd}}-s\omega_z$, $s=-\mathrm{sign}(g_z)$, $\sigma=0.005$;
  $\hat g_{xy}^{\mathrm{ref}}$ is the FK-derived attitude, $\theta_c$
  the orientation cutoff, and $\mathbf{1}[\cdot]$ the indicator.}
  \label{tab:rewards}
  \footnotesize
  \setlength{\tabcolsep}{4pt}
  \begin{tabular}{@{}lrl@{}}
    \toprule
    Term & Coeff. & Formulation \\
    \midrule
    \multicolumn{3}{@{}l}{\emph{Tracking and task (positive)}} \\
    tracking\_lin\_vel\_x   & $+3.0$  & $\exp(-e_v^2/\sigma)$ \\
    tracking\_ang\_vel      & $+1.5$  & $\exp(-e_\omega^2/\sigma)$ \\
    feet\_air\_time         & $+1.2$  & $\sum_k \max(t_k^{\text{air}} - 0.5, 0)$ \\
    orientation             & $+0.5$  & $\|\hat{g}_{xy} - \hat{g}_{xy}^{\text{ref}}\|^2 < \theta_c$ \\
    survival                & $+0.2$  & $\mathbf{1}[\text{contact}]$ \\
    \midrule
    \multicolumn{3}{@{}l}{\emph{Posture and style (negative)}} \\
    feet\_slip              & $-1.5$  & $\sum_k \|v_k^{\text{foot}}\| \cdot c_k$ \\
    feet\_toward\_gravity   & $-1.0$  & $\sum_k (1 - \hat{f}_k \cdot \hat{g})$ \\
    default\_pose           & $-0.2$  & $\sum_i \text{wrap}(\theta_i - \theta_i^{\text{ref}})^2$ \\
    \midrule
    \multicolumn{3}{@{}l}{\emph{Smoothness and effort (negative)}} \\
    torques                 & $-10$   & $\sum_i \tau_i^2$ \\
    action\_rate            & $-1.0$  & $\sum_i (a_i^t - a_i^{t-1})^2$ \\
    ang\_vel\_xy            & $-0.5$  & $\omega_x^2 + \omega_y^2$ \\
    vel\_tracking\_error    & $-0.5$  & $e_v^2 + e_\omega^2$ \\
    dof\_acc                & $-3\!\cdot\!10^{-7}$ & $\sum_i \ddot{\theta}_i^2$ \\
    dof\_vel                & $-4\!\cdot\!10^{-5}$ & $\sum_i \dot{\theta}_i^2$ \\
    \midrule
    \multicolumn{3}{@{}l}{\emph{Safety and soft limits (negative)}} \\
    collision               & $-10$   & $\mathbf{1}[\text{non-foot contact}]$ \\
    pgz\_crossings          & $-10$   & $\mathbf{1}[\text{sign}(g_z^t) \neq \text{sign}(g_z^{t-1})]$ \\
    torque\_limits          & $-1.0$  & $\max(|\tau_i| - 0.6\tau_{\max}, 0)^2$ \\
    dof\_vel\_limits        & $-0.5$  & $\max(|\dot\theta_i| - 0.9\dot\theta_{\max}, 0)^2$ \\
    \bottomrule
  \end{tabular}
\end{table}

\subsection{Hardware Measurement and Evaluation}
All hardware experiments use the same trained policy on one MiNI-Q assembly.
An external ZED2 camera tracks AprilTags for trajectory measurement and
supplies no policy input. Sim-to-sim validation uses
MuJoCo~\citep{todorov2012mujoco} at $\mathrm{d}t=\SI{1}{\milli\second}$
in effort mode and Isaac Lab at $\mathrm{d}t=\SI{2}{\milli\second}$ in
position mode, both under the identified actuator parameters. The
tracking comparison includes frozen-policy rollouts in both engines.
Hardware recovery trials take place on a
laminated tabletop; foot swaps are tested on MDF, polycarbonate, turf,
and rubber, with reported static friction coefficients from $0.3$ to
$0.7$. These surfaces support controlled evaluation for the
platform's intended laboratory and desktop use.

Trial counts balance per-condition reporting against mechanical wear
from high-energy releases: velocity tracking uses $N=5$ per
orientation; the \SI{25}{\second} posture trial is repeated five
times; recovery uses $N=6$ per condition across five conditions
(30 trials); foot variants use $N=3$ per geometry--surface cell.
Each experiment reports its success or tracking criterion alongside
the results. Error bars and shaded bands denote one standard
deviation. Accompanying video for every experiment is available on the
project website listed in the abstract.

\section{Results}
\label{sec:experiments}

Hardware tests share one unchanged checkpoint across orientations,
postures, and feet; simulation ablations retrain each design variant.

% Posture tracking is queued in setup; the remaining figures follow here.

\begin{figure*}[t]
 \centering
 \begin{minipage}[c]{0.39\textwidth}
  \includegraphics[width=\linewidth]{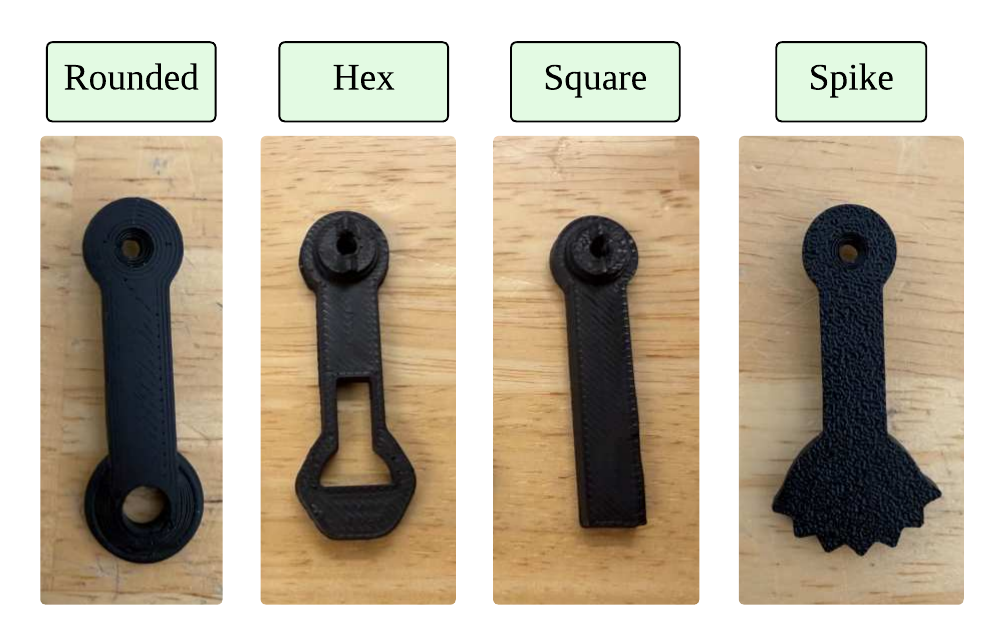}
 \end{minipage}\hfill
 \begin{minipage}[c]{0.58\textwidth}
  \centering
  \small
  \begin{tabular}{@{}lcccc@{}}
   \toprule
   Geometry & MDF & Polycarbonate & Turf & Rubber \\
   Reported $\mu_s$ & 0.3 & 0.4 & 0.5 & 0.7 \\
   \midrule
   Rounded & \textbf{0.041} & \textbf{0.065} & \textbf{0.060} & \textbf{0.064} \\
   Hex & 0.055 & 0.096 & 0.098 & 0.116 \\
   Square & 0.072 & 0.114 & 0.088 & 0.087 \\
   Spike & 0.075 & 0.078 & 0.136 & 0.065 \\
   \bottomrule
  \end{tabular}
 \end{minipage}
 \caption{Foot-geometry transfer on hardware. Left: interchangeable
 printed variants. Right: forward-velocity RMSE (m/s), $N=3$ per cell.
 Rounded is represented in training; Hex, Square, and Spike are held-out
 physical shapes. Bold marks the lowest error on each surface.}
 \label{fig:foot_dr}
\end{figure*}

\begin{figure*}[t]
 \centering
 \begin{minipage}[c]{0.20\textwidth}
  \includegraphics[width=\linewidth]{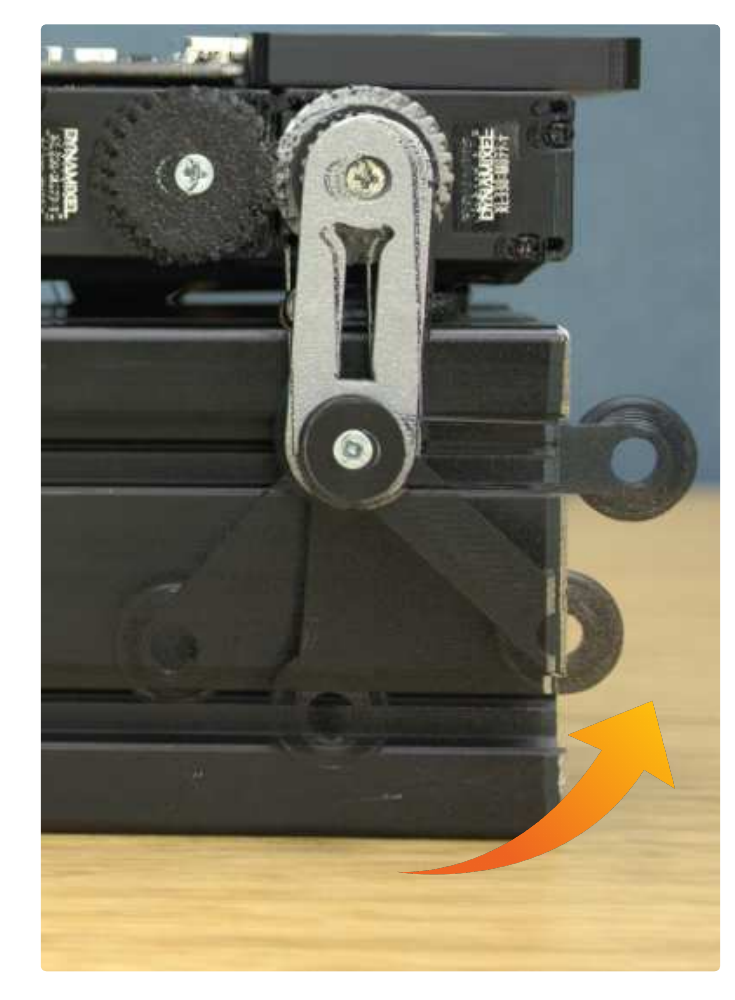}
 \end{minipage}\hfill
 \begin{minipage}[c]{0.76\textwidth}
  \includegraphics[width=\linewidth]{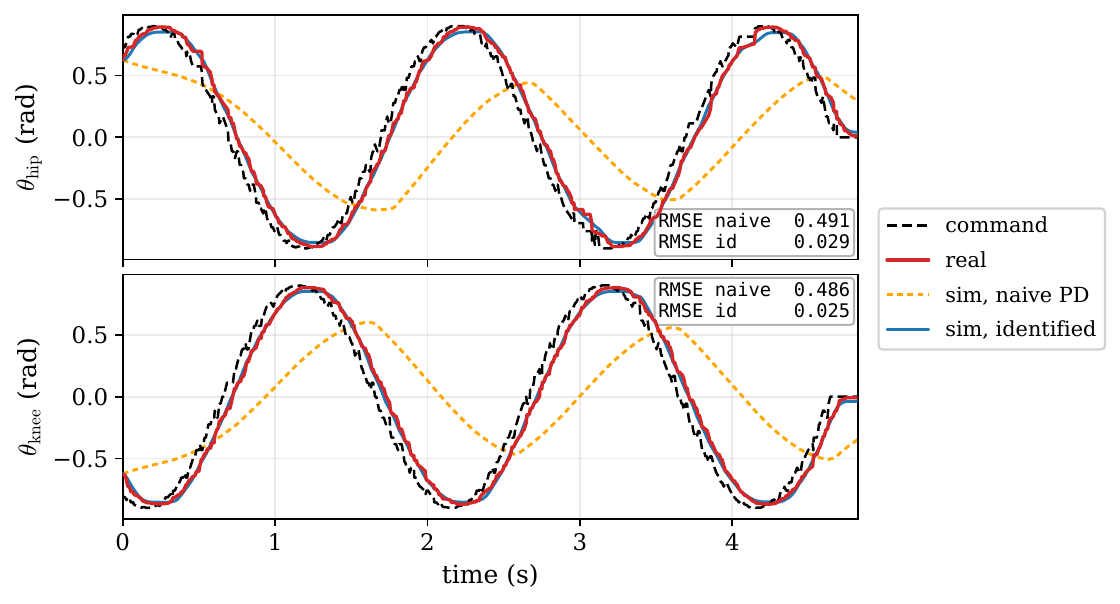}
 \end{minipage}
 \caption{Actuator identification. Left: single-leg fixture. Right:
 hip and knee trajectories for the command (dashed), measured motion
 (red), nominal-gain PD (orange), and fitted model (blue).}
 \label{fig:sysid}
\end{figure*}

\begin{figure*}[t]
 \centering
 \includegraphics[width=\textwidth]{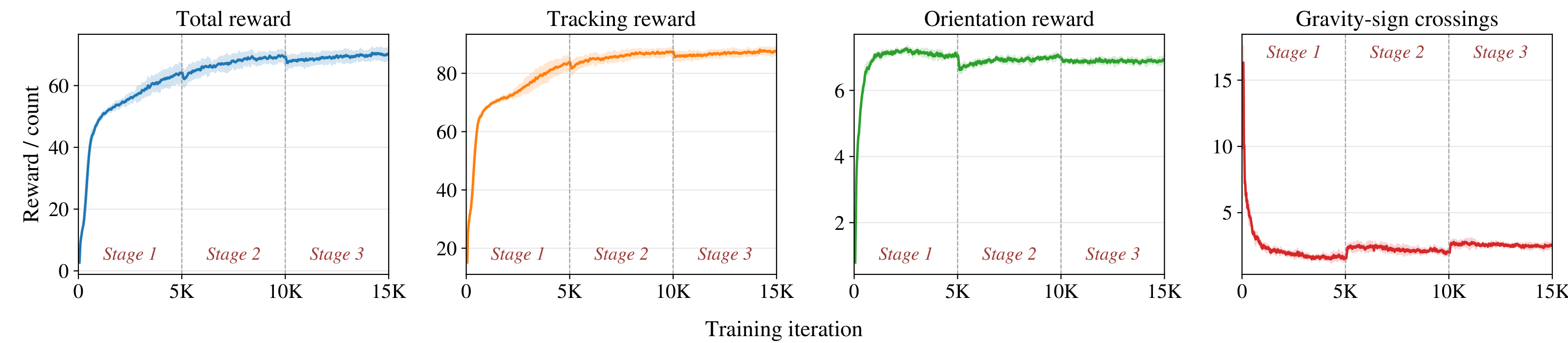}
 \caption{Full-policy training diagnostics, mean $\pm1\sigma$ over four
 seeds. Panels report total, tracking, and orientation rewards and
 gravity-sign crossings. Boundaries mark curriculum transitions at 5k
 and 10k; the first 15k of 20k iterations are shown.}
 \label{fig:learning}
\end{figure*}

\subsection{Upright and Inverted Velocity Tracking}
\label{sec:exp_velocity}
\label{sec:exp_z2}
Figure~\ref{fig:tracking} compares the same forward-velocity and yaw-rate
ramp-and-hold commands in upright and inverted orientations. Hardware
forward-velocity RMSE is $0.037\pm0.004$~m/s upright and
$0.050\pm0.003$~m/s inverted over five trials per orientation.
At the evaluated \SI{0.15}{\meter\per\second} command, these errors
are approximately $25\%$ and $33\%$: the shared controller sustains both
orientations with lower precision when inverted. The upright and
inverted panels in Figure~\ref{fig:tracking} are visually close,
indicating that the gravity-conditioned reference does not require
retraining or a separate evaluation protocol.

Yaw tracking exhibits overshoot and oscillation at motion onset before
settling. Without a hip abduction--adduction actuator, the robot
cannot generate yaw from a standstill without a large initial
excursion in $T^8$ to break static friction and initiate rotation;
once moving, the policy recovers and tracks the commanded yaw rate.
A trot emerges without an explicit gait schedule, and lateral
drift varies by surface, a consequence of the missing hip
abduction--adduction degree of freedom. Projected-gravity components
remain within $|g_x|,|g_y|<0.08$ in these tracking trials. Turning
onset and lateral stability are consequently targets for controller
and hardware refinement.

\subsection{Posture Changes During Walking}
\label{sec:exp_morphology}
Figure~\ref{fig:pose_tracking} shows a continuous \SI{25}{\second} trial
that commands ANYmal, Nose-down, Nose-up, Crouch, and Q8bot references
while maintaining a forward-velocity command. The trial begins in
ANYmal ($d_{T^8}\approx 0$), interpolates through nearby attitudes
around $t=5$~s, commits to Crouch ($d_{T^8}\approx 2.0$~rad) at
$t\approx 13$~s, and finally to Q8bot ($d_{T^8}\approx 3.0$~rad).
Each transition appears as a step in $d_{T^8}$ followed by settling
to the target band. Over five repetitions, mean forward-velocity RMSE
is approximately \SI{0.04}{\meter\per\second}. Brief dips at
transitions reflect the body reorienting before the trot resumes.
The robot changes its leg configuration and resumes tracking without
changing policy weights.

The images and $d_{T^8}$ trace show configuration changes alongside
continued locomotion. The distance is measured to the fixed ANYmal
anchor, not the active target. This experiment exercises stance and
attitude commands from the trained anchor/interpolation family through
the same interface.

\subsection{Recovery After Release}
\label{sec:exp_throw}
A recovery succeeds if the robot reaches stable commanded walking
within \SI{5}{\second} of first contact. Hardware releases use heights
of \SIrange{20}{30}{\centi\meter} to limit impact-related hardware risk,
below the \SI{0.5}{\meter} training ceiling. Table~\ref{tab:recovery} reports
24 successes in 30 trials across upright and inverted pitch and roll
perturbations and a throw condition. The same policy remains active
through the release and landing.

\begin{table}[t]
 \centering
 \caption{Hardware release recovery: stable walking within
 \SI{5}{\second} of first contact. All failures count in the outcome.}
 \label{tab:recovery}
 \footnotesize
 \begin{tabular}{@{}lrr@{}}
  \toprule
  Condition & Successes / trials & Success rate \\
  \midrule
  Upright, pitch & 4/6 & 67\% \\
  Inverted, pitch & 4/6 & 67\% \\
  Upright, roll & 5/6 & 83\% \\
  Inverted, roll & 6/6 & 100\% \\
  Throw & 5/6 & 83\% \\
  \midrule
  Overall & 24/30 & 80\% \\
  \bottomrule
 \end{tabular}
\end{table}

The six failures include actuator overcurrent shutoff, in which the
actuator latches into a no-torque state and the robot collapses;
battery-contact dislodgement; and one belt-tooth slip that misaligned
a leg. These events identify actuator impact tolerance, power
continuity, and transmission alignment as reliability priorities
rather than loss of closed-loop control; all count in
Table~\ref{tab:recovery}. Per-trial sequences are recorded in
Appendix~\ref{app:throw_breakdown}.

\subsection{Transfer Across Replaceable Foot Geometries}
\label{sec:exp_foot}
The nominal foot insert has an effective diameter of approximately
\SIrange{12}{16}{\milli\meter} under manufacturing and wear variation,
motivating the randomized geometry and contact offset. We replace the
rounded lower-link geometry with hexagonal, square, and
spike variants absent from the training asset pool. The same policy is
used without retraining, adaptation, or a foot-shape input.
Figure~\ref{fig:foot_dr} reports forward-velocity RMSE at a command of
\SI{0.15}{\meter\per\second} with zero commanded yaw, using three
trials per geometry--surface combination.

The system walks in all 16 tested combinations. Rounded feet have the
lowest error on each surface, from $0.041$ on MDF to $0.065$ on
polycarbonate. Across the 12 unseen-shape cells the RMSE remains under
$0.14$~m/s, with median $0.087$ and range
\SIrange{0.055}{0.136}{\meter\per\second}. The largest error occurs for
Spike on turf, where the elongated tip visibly catches on fibers; the
same Spike foot on the more rigid surfaces stays below $0.08$. The
error variation makes geometry--surface compatibility an engineering
constraint even when one policy supports every tested combination.
A matched sphere-only hardware comparison would isolate the benefit of
foot randomization within this transferred system.

\subsection{Actuator Fidelity and Embedded Execution}
\label{sec:exp_system}
Figure~\ref{fig:sysid} shows the identification fixture and measured
versus simulated joint trajectories. The nominal-gain PD baseline
omits armature and friction terms. The fitted model reduces the
approximately \SI{0.49}{\radian} RMSE of the nominal-gain PD baseline to
\SI{0.029}{\radian} at the hip and \SI{0.025}{\radian} at the knee, a
$17$--$20\times$ reduction.
Evaluation on a held-out XL330 gives \SI{0.019}{\radian} RMSE.
Table~\ref{tab:sysid} gives the fitted model, including the
\SI{0.228}{\newton\meter} torque bound.

\begin{table}[t]
 \centering
 \caption{Effective actuator-model settings used for training and
 evaluation. $\tau_{\max}$ specifies the simulated torque bound.}
 \label{tab:sysid}
 \footnotesize
 \begin{tabular}{@{}llr@{}}
 \toprule
 Symbol & Parameter & Value (SI units) \\
 \midrule
 $k_p$ & Position stiffness & $5.95\times10^{-1}$ \\
 $k_d$ & Position damping & $2.74\times10^{-2}$ \\
 $b$ & Viscous damping & $1.01\times10^{-4}$ \\
 $I_a$ & Rotor armature & $6.57\times10^{-4}$ \\
 $\tau_f$ & Friction loss & $2.86\times10^{-2}$ \\
 $\tau_{\max}$ & Simulated torque bound & $2.28\times10^{-1}$ \\
 \bottomrule
 \end{tabular}
\end{table}

The lightweight links also exposed sensitivity to effort integration
and contact handling during transfer. Both training and deployment
use position targets.

Table~\ref{tab:nn_perf} reports onboard inference time and memory across
three candidate architectures. The deployed $[128,128]$ policy takes
\SI{594.33}{\micro\second}, approximately $3\%$ of the
\SI{20}{\milli\second} policy period, compared with approximately
\SI{4000}{\micro\second} for the scalar implementation. The reported
implementation footprint is \SI{150.3}{\kilo\byte} RAM and
\SI{846.7}{\kilo\byte} flash. Keeping weights in DRAM avoids
flash-access overhead during the repeated matrix--vector products.
This leaves execution headroom for observation construction and
sensor/actuator exchange; the benchmark measures the forward pass,
excluding sensor and bus latency.

\begin{table}[t]
 \centering
 \caption{Implementation memory footprint and forward-pass latency
 for candidate policy sizes. Bold marks the deployed configuration.}
 \label{tab:nn_perf}
 \footnotesize
 \begin{tabular}{@{}lrrr@{}}
  \toprule
  Hidden layers & RAM (KB) & Flash (KB) & Inference ($\mu$s) \\
  \midrule
  $128\times64$ & 115.8 & 812.4 & 400.74 \\
  $\mathbf{128\times128}$ & \textbf{150.3} & \textbf{846.7} & \textbf{594.33} \\
  $256\times128$ & 247.3 & 943.2 & 1154.55 \\
  \bottomrule
 \end{tabular}
\end{table}

\subsection{Training Across the Curriculum}
\label{sec:exp_training}
Figure~\ref{fig:learning} shows four-seed training diagnostics.
Orientation reward briefly dips when the release distribution broadens
at 5k and 10k iterations, then recovers; gravity-sign crossings
(the count of projected-gravity $z$ sign changes within an episode)
decline as the policy learns to settle into an orientation rather
than oscillate across the upright--inverted threshold. The curves
level off within the first 15k iterations shown from the 20k schedule
(Table~\ref{tab:curriculum}), which is evidence that the unified
policy is learned from scratch in one pass without warm-start or
modular composition. This supports progressively broadening
the posture/release distribution while retaining one policy. The
following ablations use one training run per variant.

\subsection{Ablation of the Control Formulation}
\label{sec:ablations}
Table~\ref{tab:ablations} compares the full design with five modified
training runs. Evaluation uses 200 parallel environments per
anchor--command cell, averaged over Crouch, ANYmal, and Q8bot, with one
training run per design variant. Tracking fractions are achieved over
commanded velocity; recovery time
is capped at \SI{10}{\second} after releases from \SI{0.35}{\meter} with
$\mathrm{SO}(3)$ orientations. This protocol differs from the hardware
five-second success criterion in Table~\ref{tab:recovery}.
Appendix~\ref{app:ablations} defines the five interventions.

\begin{table}[t]
 \centering
 \caption{Simulation ablations. Tracking columns report achieved / commanded
 velocity ($u$: upright; $i$: inverted); recovery time is in seconds.}
 \label{tab:ablations}
 \footnotesize
 \setlength{\tabcolsep}{4pt}
 \begin{tabular}{@{}lrrrr@{}}
  \toprule
  Design & $v_x^u$ & $v_x^i$ & $\omega_z$ & Recovery \\
  \midrule
  \textit{Unified Baseline} & \textbf{0.98} & \textbf{1.00} & \textbf{0.79} & \textbf{6.9} \\
  No toroidal encoding & 0.78 & 0.21 & 0.61 & 9.7 \\
  No $\mathbb{Z}_2$ mirror & 0.59 & 0.02 & 0.27 & 10.0 \\
  No throw curriculum & 0.76 & 0.80 & 0.77 & 9.7 \\
  Discrete morphology & 0.69 & 0.00 & 0.39 & 10.0 \\
  No foot DR & 0.96 & 0.94 & 0.76 & 8.4 \\
  \bottomrule
 \end{tabular}
\end{table}

Removing the gravity-conditioned reference reduces inverted tracking
from $1.00$ to $0.02$, while replacing periodic features with raw angles
reduces it to $0.21$. Restricting training to phase-1 releases increases
mean recovery time from \SI{6.9}{\second} to \SI{9.7}{\second}.
The discrete-posture variant also fails to sustain inverted walking,
but changes both conditioning and interpolation training; its result
does not establish that continuous encoding is necessary in general.

Without foot randomization, sphere-foot tracking remains near baseline
and recovery slows to \SI{8.4}{\second}. The hardware swaps demonstrate
robustness of the full policy across contact shapes, but a matched
sphere-only hardware policy is needed to isolate the benefit of
randomization. These comparisons characterize the tested variants;
one training run per variant does not quantify training-seed variability.

\section{Discussion}
\label{sec:limitations}

Continuous rotation gives one policy access to upright and inverted
supporting configurations, so recovery can end in walking on either
side of the body. The reference transformation and posture commands
connect these behaviors without a separate recovery controller.
The hardware experiments demonstrate this shared interface across
postures, landings, and interchangeable feet.

Evaluation covers one assembly, four indoor surfaces, the selected
postures, and \SIrange{20}{30}{\centi\meter} releases.
Cross-unit variability, outdoor terrain, and large initial angular
velocities remain uncharacterized. Tracking errors are substantial
relative to the \SI{0.15}{\meter\per\second} command; release failures
include overcurrent shutoff, battery dislodgement, and belt slippage.
Anchor coverage is also limited by training stability when distant
posture pairs are introduced together
(Appendix~\ref{app:anchor_selection}).

Repeated-seed ablations, cross-unit tests, and direct attitude tracking
would strengthen the evaluation. Matched specialist and sphere-only
hardware policies would clarify the costs of unification and the benefit
of foot randomization. Beyond these tests, selecting postures according
to task and contact conditions would extend the current interface
beyond prescribed commands.

\section{Conclusion}
\label{sec:conclusion}

We demonstrated a single learned policy running entirely on the
onboard microcontrollers of a \SI{270}{\gram} quadruped whose
belt-driven legs rotate continuously through \SI{360}{\degree}.
To our knowledge, this is the first onboard learned policy at the
sub-kilogram scale that combines throw-and-walk recovery,
multi-posture conditioning, and foot-shape generalization in one
network. Continuous posture references on $T^8$ and the
leg-symmetry transformation enable upright and inverted walking,
posture changes, and landing recovery; calibrated actuation,
cross-engine validation, and foot randomization support transfer
across interchangeable feet. The policy runs at \SI{50}{\hertz} on a
dual-microcontroller stack. Together, these results show how
continuous-rotation embodiment can expand the behaviors available to
a miniature robot under one controller. This capability opens a path
to task-driven posture selection and collaborative tasks in which
multiple miniature robots coordinate their postures and contact
choices.

\section*{Acknowledgments}

This work was supported in part by the Amazon AI PhD Fellowship
awarded to Arturo Flores Alvarez.

% Keep normal column flow through the bibliography before the appendix.
\nobalance
\bibliographystyle{IEEEtranN}
\bibliography{miniq}

\clearpage
\nobalance
\appendices
% Supporting material. Tables and figures that already appear in the
% combined main PDF are omitted here when \mainpaper is defined.
\section{Engineering Details}
\label{app:engineering}
\label{app:sysid}

\subsection{XL330 Identification Procedure}
\label{app:sysid_details}
The actuator-identification experiment mounts a single leg on a floating
fixture with the base free along gravity and the end effector
unconstrained. Each joint is excited independently using step references
and a low-frequency sinusoidal sweep at the firmware nominal gains
$P=D=400$. MuJoCo runs at \SI{1}{\kilo\hertz} in effort mode under
explicit torque commands, hard-clipped at the actuator torque bound.
We fit effective stiffness $k_p$, damping $k_d$, armature $I_a$, viscous
damping $b$, and friction loss $\tau_f$ by gradient-free Nelder--Mead
minimization of joint-position and joint-velocity trajectory error. A
narrow stiffness constraint around the firmware nominal reduces the
$k_p$--armature degeneracy that appears once the commanded effort
saturates.

The identified parameters transfer between hip and knee joints and are
then reused across MuJoCo, Isaac Gym, and Isaac Lab under the appropriate
drive interface. Table~\ref{tab:sysid} lists the fitted values. As
reported in Figure~\ref{fig:sysid}, the fitted model reduces the
approximately \SI{0.49}{\radian} nominal-gain PD error to
\SI{0.029}{\radian} at the hip and \SI{0.025}{\radian} at the knee; a
held-out XL330 unit gives \SI{0.019}{\radian} RMSE. The nominal baseline
uses the datasheet/firmware gains converted to physical units without
armature or friction terms.

\ifdefined\mainpaper
\else
\begin{table}[htbp]
 \centering
 \caption{Effective XL330-M077-T actuator-model parameters.
 $\tau_{\max}$ specifies the simulated torque bound.}
 \label{tab:sysid}
 \begin{tabular}{lll}
 \toprule
 Symbol & Parameter & Value (SI units) \\
 \midrule
 $k_p$ & Position stiffness & $5.95\times10^{-1}$ \\
 $k_d$ & Position damping & $2.74\times10^{-2}$ \\
 $b$ & Viscous damping & $1.01\times10^{-4}$ \\
 $I_a$ & Rotor armature & $6.57\times10^{-4}$ \\
 $\tau_f$ & Friction loss & $2.86\times10^{-2}$ \\
 $\tau_{\max}$ & Simulated torque bound & $2.28\times10^{-1}$ \\
 \bottomrule
 \end{tabular}
\end{table}
\fi

\subsection{Simulation Configuration and Interface Checks}
\label{app:simulation}
\label{app:sim2sim}
Table~\ref{tab:sim_config} records the physics steps and drive interfaces
used for training, actuator identification, and evaluation. The fitted
actuator parameters are shared across these configurations.

\begin{table}[htbp]
 \centering
 \caption{Physics integration and drive interfaces. All engines use
 the identified actuator model and a \SI{50}{\hertz} actor.}
 \label{tab:sim_config}
 \footnotesize
 \setlength{\tabcolsep}{4pt}
 \begin{tabular}{@{}llll@{}}
 \toprule
 Engine & Step & Drive & Role \\
 \midrule
 Isaac Gym & 2 ms & Position & Training \\
 MuJoCo & 1 ms & Explicit effort & Identification, evaluation \\
 Isaac Lab & 2 ms & Position & Evaluation \\
 \bottomrule
 \end{tabular}
\end{table}

We validate parameter transfer by replaying a frozen rollout across
MuJoCo, Isaac Gym, and Isaac Lab under the same actor and identified
parameters. Observation-pipeline equivalence is checked separately by
feeding the same recorded joint and base states through each engine's
observation builder; the resulting vectors agree to the
single-precision floor, checking ordering, scaling, and angle
conventions independently of the physics.

Closed-loop evaluation exposed a hard failure specific to Isaac Gym's
effort regime for the approximately \SI{1}{\gram} lower links. At the
tested \SI{2}{\milli\second} step, a commanded torque is held across
multiple solver iterations and interacts with the contact solver
differently from MuJoCo's atomic \SI{1}{\milli\second} step. The
cross-engine replay caught this integration artifact before hardware
deployment. The final checkpoint is therefore trained with position
drives in Isaac Gym, evaluated with explicit effort in MuJoCo and
position drives in Isaac Lab, and deployed through the robot's native
position interface.

\subsection{Dual-Microcontroller Onboard Inference}
\label{app:dualmcu}
The compiled actor runs on a dedicated S3 core as a high-priority
FreeRTOS task at \SI{50}{\hertz}; the second core services the
I\textsuperscript{2}C link to the C3 and any non-control-loop activity
(Figure~\ref{fig:realtime}). This partitioning gives deterministic
execution without interference from peripheral traffic. The forward
pass uses the S3's PIE128 SIMD instructions, vectorizing the
matrix--vector products that dominate runtime and reducing inference
latency from approximately \SI{4000}{\micro\second} (scalar) to under
\SI{600}{\micro\second}. Model weights are copied from flash into DRAM
at startup to eliminate flash-access overhead and to improve cache
locality for the small layer sizes. Table~\ref{tab:nn_perf} reports
the per-architecture benchmark; the chosen $[128,128]$ configuration
leaves headroom for larger observation vectors or additional task
heads.

\section{Representation and Training}
\label{app:representation_training}

\subsection{Periodic Representation and Anchor Geometry}
\label{app:toroidal_math}
\label{app:anchor_selection}

\begin{figure*}[t]
 \centering
 \includegraphics[width=\textwidth]{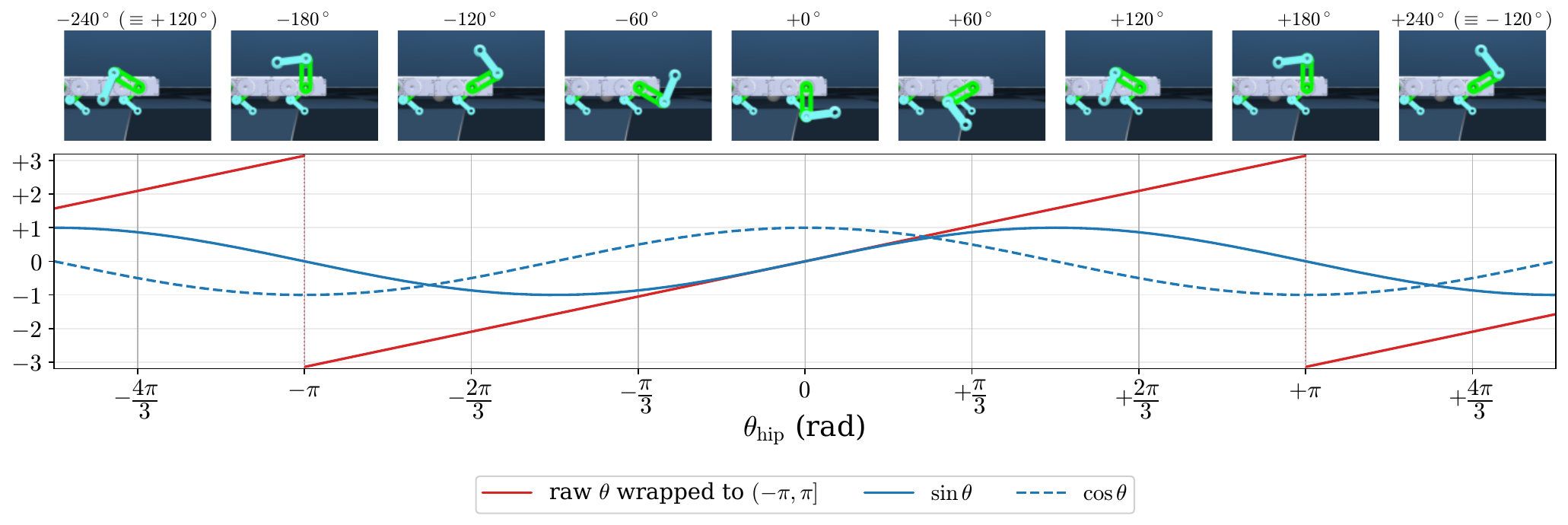}
 \caption{Periodic joint representation. Top: the front-right hip at
 nine positions covering $\pm 4\pi/3$; poses at $\pm 180^{\circ}$ and
 $\pm 240^{\circ}$ ($\equiv\mp 120^{\circ}$) coincide physically but
 differ by $2\pi$ in raw $\theta$. Bottom: raw $\theta$ wrapped to
 $(-\pi,\pi]$ jumps at every odd multiple of $\pi$; the
 $(\sin\theta,\cos\theta)$ encoding used by the policy remains smooth.}
 \label{fig:leg_circle}
\end{figure*}

Figure~\ref{fig:leg_circle} illustrates the angle-wrap boundary and the
sine/cosine representation. The map
$\theta\mapsto(\sin\theta,\cos\theta)$ is one-to-one and smooth on all
of $S^1$. If two angles have geodesic separation $\delta\in[0,\pi]$,
their embedding distance is $2\sin(\delta/2)$.
Consequently, $(2/\pi)\delta\leq2\sin(\delta/2)\leq\delta$.
Applying this independently to eight joints gives a bounded-distortion
embedding of $T^8$ in $\mathbb{R}^{16}$, which is unattainable under
raw-angle parameterization. The bound relates separation in the
periodic input features to separation in joint configuration.
The gravity-conditioned reference is a separate, sign-dependent
selection applied before constructing these features.

The seven anchor postures shown in the main paper contain three stance
configurations and four attitude perturbations. Figure~\ref{fig:anchor_heatmap}
reports their pairwise toroidal distances. The set is chosen so that
exactly one pair lies above the stage-2 nearby-pair threshold
$r_2=\SI{4.0}{\radian}$, avoiding the pattern in which multiple far
pairs enter stage~3 simultaneously. ANYmal sits at the centroid of the
anchor cloud: the attitude anchors are within \SI{1.6}{\radian} of it,
while Crouch and Q8bot are \SI{2.86}{\radian} and \SI{3.16}{\radian}
from that reference. The Lean-left--Q8bot pair has distance
\SI{4.03}{\radian}, just beyond the nearby-pair curriculum threshold,
and is the single stage-3 graduation pair. The selection was informed
by preliminary training attempts in which admitting multiple distant
pairs together destabilized training.

\ifdefined\mainpaper
\else
\begin{figure}[htbp]
 \centering
 \includegraphics[width=0.65\textwidth]{10_appendix/anchor_heatmap_matrix.pdf}
 \caption{Pairwise toroidal distances among the training reference
 postures, used to define the nearby-pair curriculum.}
 \label{fig:anchor_heatmap}
\end{figure}
\fi

\subsection{Attitude Control as a Derived Capability}
\label{app:attitude_control}
Four of the seven training anchors (Nose-up, Nose-down, Lean-left,
Lean-right) command non-zero body pitch or roll. When the policy is
conditioned on one of these, it holds the corresponding tilted
attitude while walking. The behavior emerges from the default-pose and
orientation reward terms applied to the anchor: planar two-degree-of-freedom
leg forward kinematics maps each anchor's joint angles to an expected
body attitude in projected-gravity coordinates, so a tilted anchor
produces a tilted target without a reward designed specifically for
attitude control. Hardware demonstrations of the four attitude
commands are included on the project website listed in the abstract.

\subsection{Privileged Critic Observation}
\label{app:priv_obs}
The 22-dimensional privileged observation supplied to the critic, and
not to the actor, contains: base linear velocity (3), base height
above terrain (1), applied push forces and torques from the
disturbance randomizer ($3+3$), noise-free joint velocities from
PhysX (8), and a per-environment base-mass scaling factor (4). These
quantities are either unavailable or substantially noisier on the real
robot and are therefore excluded from the deployment observation.

\subsection{Reward and Training Configuration}
\label{app:tables}
\label{app:rewards}
\label{app:hyperparams}

Table~\ref{tab:rewards} lists the reward configuration, including the
posture-dependent attitude reward. In the expressions, $c_k$ denotes foot contact, $\hat f_k$ the
foot direction used by the gravity-alignment term, and $t_k^{\mathrm{air}}$
the corresponding swing duration. Table~\ref{tab:hyperparams} gives the
training settings. The main paper contains the release/posture curriculum
and the complete deployment observation.

\ifdefined\mainpaper
\else

\fi

\begin{table}[!htbp]
    \centering
    \caption{PPO and training hyperparameters. The entropy term enters an objective that is minimized, so the negative coefficient corresponds to a conventional entropy bonus.}
    \label{tab:hyperparams}
    \small
    \begin{tabular}{ll}
      \toprule
      Hyperparameter & Value \\
      \midrule
      Parallel environments      & 4096 \\
      Rollout horizon            & 24 steps \\
      PPO mini-epochs            & 20 per rollout \\
      KL target (adaptive LR)    & $\approx 0.01$ \\
      Entropy coefficient        & $-0.015$ \\
      Discount $\gamma$          & 0.99 \\
      GAE $\lambda$              & 0.95 \\
      Total iterations           & 20{,}000 \\
      Control rate               & \SI{50}{\hertz} \\
      Actor / critic             & $[128, 128]$ ELU MLP \\
      \bottomrule
    \end{tabular}
\end{table}

The critic receives 22 privileged channels describing base linear
velocity and height, applied force and torque disturbances, noise-free
joint velocities, and base mass properties
(Appendix~\ref{app:priv_obs}). The deployed actor receives
none of these privileged channels. Training runs at roughly
70k environment steps per second on one NVIDIA RTX 4090, with the
20k-iteration schedule taking approximately eight hours.

Figure~\ref{fig:learning} shows the training diagnostics across four
seeds for the full configuration. Channels report total episode
reward, tracking reward (linear and angular velocity terms),
orientation reward, and gravity-sign crossings per episode. A
gravity-sign crossing is a change in the sign of projected-gravity
$z$ within an episode, so the count serves as a proxy for repeated
oscillation across the upright--inverted boundary. The curves show the
first 15k iterations of the 20k schedule because all four channels have
largely flattened by that point.

Curriculum boundaries at 5k (lateral drops) and 10k (full
$\mathrm{SO}(3)$ spawn) appear as transient dips in orientation reward
followed by recovery, while gravity-sign crossings fall as the policy
learns to settle into a supporting orientation. All runs begin from
random initialization: the unified controller is learned in one pass
without warm-start, pre-training, or modular policy composition. The
full configuration trains at roughly 70k environment steps per second
on one NVIDIA RTX 4090 (approximately 100k steps/s for baseline walking
without the full feature set), with the complete 20k-iteration schedule
taking approximately eight hours. The main-paper ablations use one
training run per modified design.

\ifdefined\mainpaper
\else
\begin{figure}[htbp]
 \centering
 \includegraphics[width=\textwidth]{10_appendix/learning.pdf}
 \caption{Training diagnostics for the full configuration, mean
 $\pm1\sigma$ over four seeds. Channels report total reward, tracking
 reward, orientation reward, and gravity-sign crossings per episode.
 Curriculum transitions occur at iterations 5k and 10k.}
 \label{fig:learning}
\end{figure}
\fi

\section{Evaluation Details}
\label{app:evaluation}

\subsection{Ablation Interventions}
\label{app:ablations}
\label{app:ablation_summary}
\label{app:ablation_interventions}

Each row of Table~\ref{tab:ablations} corresponds to a retrained variant
of the full controller. The interventions below match the table order;
the discrete-morphology variant changes two coupled design choices.
Section~\ref{sec:ablations} gives the common evaluation protocol and
interprets the results.

\noindent\textbf{No Toroidal Encoding.}
The variant replaces $(\sin\theta,\cos\theta)$ with the raw joint angle
$\theta$ in both the joint-offset observations and the posture-command
input. It tests the periodic representation across both inputs.

\noindent\textbf{No \texorpdfstring{$\mathbb{Z}_2$}{Z2} Mirror.}
The variant disables the gravity-conditioned reference flip, retaining
the upright reference regardless of projected-gravity sign.

\noindent\textbf{No Throw Curriculum.}
The variant freezes the release distribution at the phase-1 setting
throughout training. It therefore removes exposure to later release
conditions as well as the progression between stages; it does not
compare a curriculum against training on the final distribution from
the outset.

\noindent\textbf{Discrete Morphology.}
The variant replaces the continuous 16-channel reference encoding with
seven one-hot anchor labels and disables interpolation between anchors.
It jointly tests the conditioning representation and posture-training
distribution. Its failure to sustain inverted walking does not rule out
other discrete command sets that explicitly represent inverted targets.

\noindent\textbf{No Foot Domain Randomization.}
The variant trains with the nominal sphere foot only. Since the
simulation evaluation also uses sphere feet, this comparison measures
performance on that geometry. Hardware swaps in
Section~\ref{sec:exp_foot} evaluate the full policy on additional shapes;
without a matched hardware comparison, they do not isolate the effect
of randomization.

\subsection{Per-Trial Recovery Outcomes}
\label{app:throw_breakdown}

Table~\ref{tab:recovery_trials} preserves the recorded binary outcome sequence
for each hardware release condition. These are the same 30 trials
summarized in the main paper, with the same five-second success
criterion.

\begin{table}[!htbp]
    \centering
    \caption{Per-trial hardware recovery (same hardware trials as the main paper). $\bullet$ = resumed commanded locomotion, $\circ$ = did not ($N=6$ per condition).}
    \label{tab:recovery_trials}
    \small
    \begin{tabular}{@{}lcc@{}}
      \toprule
      Condition       & Trials                                      & Success \\
      \midrule
      Upright, pitch  & $\bullet\,\circ\,\circ\,\bullet\,\bullet\,\bullet$ & 4/6 \\
      Inverted, pitch & $\circ\,\bullet\,\circ\,\bullet\,\bullet\,\bullet$ & 4/6 \\
      Upright, roll   & $\bullet\,\bullet\,\bullet\,\circ\,\bullet\,\bullet$ & 5/6 \\
      Inverted, roll  & $\bullet\,\bullet\,\bullet\,\bullet\,\bullet\,\bullet$ & 6/6 \\
      Throw           & $\bullet\,\bullet\,\bullet\,\bullet\,\circ\,\bullet$ & 5/6 \\
      \midrule
      Overall         &                                             & 24/30 \\
      \bottomrule
    \end{tabular}
\end{table}

\end{document}